\documentclass[sigconf]{acmart}

\usepackage[capitalise, noabbrev]{cleveref}
\usepackage{amsmath}
\usepackage[hypcap=false]{caption}
\usepackage{subcaption}
\usepackage{multirow}
\usepackage{ctable}
\usepackage{threeparttable}
\AtBeginDocument{%
  }

\setcopyright{acmlicensed}
\copyrightyear{2018}
\acmYear{2018}
\acmDOI{XXXXXXX.XXXXXXX}

\acmConference[Conference acronym 'XX]{Make sure to enter the correct
  conference title from your rights confirmation emai}{June 03--05,
  2018}{Woodstock, NY}
\acmISBN{978-1-4503-XXXX-X/18/06}

\acmSubmissionID{5532}

\begin{document}

\title{iKAN: Global Incremental Learning with KAN for Human Activity Recognition Across Heterogeneous Datasets}


\author{Mengxi Liu}
\email{mengxi.liu@dfki.de}
\affiliation{%
  \institution{German Research Center for Artificial Intelligence}
  \streetaddress{Trippstadter Str. 122}
  \city{Kaiserslautern}
  \country{Germany}
  \postcode{67663}
}

\author{Sizhen Bian}
\email{sizhen.bian@pbl.ee.ethz.de}
\affiliation{%
  \institution{ETH Zurich}
  \streetaddress{Rämistrasse 101}
  \city{Zurich}
  \country{Switzerland}
  \postcode{8092}
}

\author{Bo Zhou}
\email{bo.zhou@dfki.de}
\affiliation{%
  \institution{German Research Center for Artificial Intelligence}
  \streetaddress{Trippstadter Str. 122}
  \city{Kaiserslautern}
  \country{Germany}
  \postcode{67663}
}

\author{Paul Lukowicz}
\email{paul.lukowicz@dfki.de}
\affiliation{%
  \institution{German Research Center for Artificial Intelligence}
  \streetaddress{Trippstadter Str. 122}
  \city{Kaiserslautern}
  \country{Germany}
  \postcode{67663}
}

\renewcommand{\shortauthors}{Authors}

\begin{abstract}

This work proposes an incremental learning (IL) framework for wearable sensor human activity recognition (HAR) that tackles two challenges simultaneously: catastrophic forgetting and non-uniform inputs.
The scalable framework, iKAN, pioneers IL with Kolmogorov-Arnold Networks (KAN) to replace multi-layer perceptrons as the classifier that leverages the local plasticity and global stability of splines.
To adapt KAN for HAR, iKAN uses task-specific feature branches and a feature redistribution layer.
Unlike existing IL methods that primarily adjust the output dimension or the number of classifier nodes to adapt to new tasks, iKAN focuses on expanding the feature extraction branches to accommodate new inputs from different sensor modalities while maintaining consistent dimensions and the number of classifier outputs. 
Continual learning across six public HAR datasets demonstrated the iKAN framework's incremental learning performance, with a last performance of 84.9\% (weighted F1 score) and an average incremental performance of 81.34\%, which significantly outperforms the two existing incremental learning methods, such as EWC (51.42\%) and experience replay (59.92\%).

\end{abstract}



\maketitle

\section{Introduction}

Comprehensive Human Activity Recognition (HAR) is a diverse and complex task that requires different sensing modalities to extract contexts ranging from sports to physiological activity.
In classical machine learning (ML), the model accesses all training data at once \cite{van2022three}. However, collecting all relevant data from all potential sensors in a single experiment is impractical in HAR scenarios.
Incrementally learning (IL) from non-stationary streams of data is a promising solution.
However, \textbf{catastrophic forgetting} and the \textbf{non-uniform input} of multimodal context resulting from modality and task heterogeneity are two major challenges of incremental learning in HAR scenarios.

A few solutions have been proposed in the past years in the ML discipline to address catastrophic forgetting \cite{masse2018alleviating,kirkpatrick2017overcoming,shin2017continual,rebuffi2017icarl}.
However, these methods mainly focus on uniform image classification tasks, which are not adaptable to HAR considering the heterogeneous nature of HAR with various modalities, channels, sampling rates, etc., across datasets.
Even in IL approaches specifically for HAR \cite{mazankiewicz2020incremental,siirtola2021context,ntalampiras2016incremental,mo2016human,adaimi2022lifelong}, the research works so far only focus on IL within each uniform dataset, and the 'new classes' are effectively masked out from the individual datasets during training process.
This is mainly due to the fact that such works are adapted from the vision domain whose architectures inherently require uniform input, as the underlying principle largely relies on expanding the classification layers or output dimensions with the same encoder for new tasks, as shown in \cref{fig: framework}.

Thus, we propose the iKAN framework specifically for IL across heterogeneous HAR datasets while addressing catastrophic forgetting.
We consider a new task using a new sensing system to detect a set of activities, which is essentially a new HAR dataset.
Unlike existing IL methods that primarily adjust the output dimension or the number of classifier nodes to adapt to new tasks, iKAN adapts task-specific feature extraction branches while keeping the dimensions of the classifier layer.
Our framework pioneers the usage of the recently proposed Kolmogorov-Arnold Networks (KAN) instead of traditional Multi-Layer Perceptrons (MLPs) as the classifier layers for IL, leveraging the local plasticity and global stability of splines\cite{liu2024kan}. 
In addition, a feature redistribution layer was designed to rescale the feature representations and leverage the properties of KAN for IL.
Unlike other HAR-specific IL methods, we tested iKAN by leaving complete datasets out as the new tasks in a single validation process (task-incremental) with six public datasets, instead of exploring within individual datasets for each validation trial (class-incremental).
Overall, we have the three following contributions:

\begin{figure*}[!t]
\footnotesize
\centering
\includegraphics[width=1.0\linewidth]{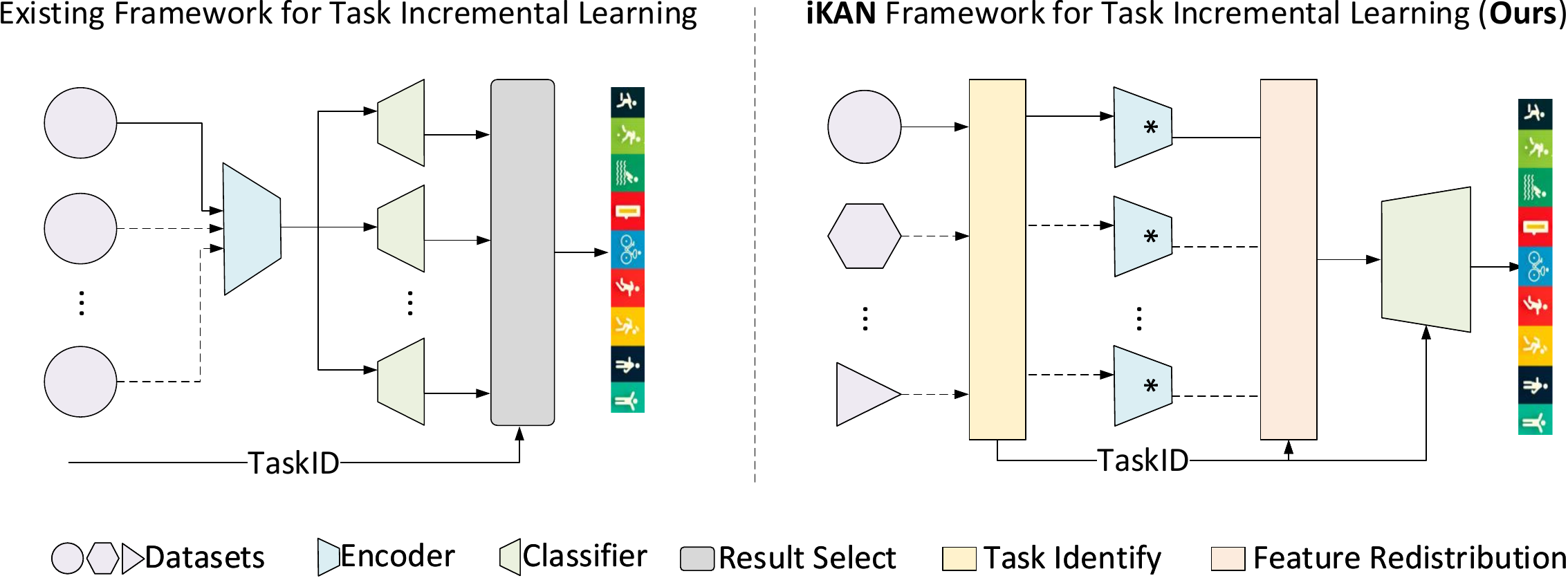}
\caption{Comparison between the existing framework for task incremental learning \cite{lu2024revisiting} and our proposed iKAN framework. (The existing framework has one encoder and increased classifiers to incrementally learn the new tasks, which can only process the task with the same input, while the iKAN framework has multiple encoders aiming to receive non-uniform inputs and one KAN-based classifier. Thus iKAN can incrementally learn the tasks across heterogeneous datasets. The shape of the datasets(tasks) represents the different sensor modalities in the datasets(tasks))}
\label{fig: framework}
\end{figure*}

\vspace{-5pt}
\begin{enumerate}
    \item A novel IL framework with a multi-encoder-single-classifier architecture boosted by KAN was proposed, enabling a single model to incrementally learn from multiple HAR datasets with minimum catastrophic forgetting (forgetting measure (0.0017), intransigence measure (0.0012)).
    \item The IL performance of iKAN was demonstrated by six public heterogeneous HAR datasets with a last performance of 84.9\% (weighted F1 score) and an average incremental performance of 81.34\% (weighted F1 score). 
    \item As the first study using KAN for IL, the optimal KAN configurations for the selected HAR tasks were identified through empirical investigation.
\end{enumerate}




\section{Related Work}
\subsection{Incremental Learning}
Enabling neural networks to incrementally learn from continuous streams of data is still an important open problem in deep learning due to catastrophic forgetting \cite{chen2018lifelong}. 
In the past few years, a number of solutions have been proposed to address the catastrophic forgetting issues in incremental learning. These include context-dependent gating (XdG), which uses context-specific components \cite{masse2018alleviating}, elastic weight consolidation (EWC), which uses parameter regularization methods \cite{liu2018rotate}, and deep generative replay (DGR) using replay methods \cite{shin2017continual}. 
While these solutions were mainly evaluated by the image datasets, incrementally learning in the sensor-based HAR scenarios is still under-explored with researchers exploring existing methods developed for computer vision in HAR \cite{adaimi2022lifelong}.
The major purpose of the existing works in HAR scenario is to develop personalized HAR models across users by adapting to distribution shifts and learning new classes from the data stream\cite{mazankiewicz2020incremental,siirtola2021context,ntalampiras2016incremental,mo2016human,adaimi2022lifelong,forster2010incremental}. 
For example, the proposed LAPNet-HAR framework by Adaimi et.el. \cite{adaimi2022lifelong}, addresses the challenge of continual learning in sensor-based human activity recognition by leveraging a prototypical network with prototype memory for incremental learning, experience replay to prevent catastrophic forgetting, continual replay-based prototype adaptation to prevent prototype obsolescence, and a contrastive loss to improve inter-class separation.
However, the existing works in the HAR scenarios only focused on incremental learning from homogeneous datasets, meaning that the structure of the input data from the new task to the incremental learning framework should remain similar, which can not continually learn new tasks requiring new sensor modalities with different input sizes like sampling rate and channels.
This situation is very common due to the diversity and complexity of human activities, which usually require different sensors to recognize different types of activities. 
Incrementally learning the new task in human activity recognition across heterogeneous datasets remains a challenge for the current framework.
To narrow this research gap, we proposed the iKAN framework, aiming to enable neural networks to continually learn new tasks independent of source characteristics in different datasets.

\subsection{Kolmogorov-Arnold Networks (KAN)}
Kolmogorov-Arnold Networks (KANs) as promising alternatives to Multi-Layer Perceptrons (MLPs) was firstly proposed in the work \cite{liu2024kan}.  KANs include learnable activation functions on edges, or "weights", whereas the MLPs have fixed activation functions on nodes, or "neurons". KANs do not use any linear weights; instead, a univariate function with spline parameterization serves as a substitute for each weight parameter.
Since spline bases are local plasticity, a new sample will only affect a few nearby spline coefficients, leaving far-away coefficients intact.
Therefore, it potentially competes with the MLP in terms of theoretical foundation, interpretability, modularity, efficiency, training stability, flexibility, and local plasticity. 
There have been some recent works that explored the advantages of the KAN in applications such as knowledge representation \cite{samadi2024smooth}, time series analysis \cite{vaca2024kolmogorov}, and pose estimation \cite{moryossef2024optimizing}.
This work, for the first time, explored the potential of KAN in incremental learning, taking the HAR application as the case study. An iKAN framework enhanced by KAN was proposed to enable the neural networks to incrementally learn new tasks across heterogeneous datasets.

\section{Framework Design}

The major purpose of the iKAN framework is to enable the neural network to incrementally learn new tasks in comprehensive HAR scenarios across heterogeneous datasets. This means that iKAN can accept new input data with different source characteristics.
A multi-encoder-single-classifier architecture was proposed in this work, standing out from the existing common architecture with single-feature-branch-multi(variant)-classifier \cite{lu2024revisiting}, where a multi-headed output layer was used, and each task had its own output units, only the output units of the task under consideration.
\cref{fig: framework} shows a comparison between the existing framework for task incremental learning and the proposed iKAN framework.
The existing framework has one encoder and increased classifiers or one classifier with variant output dimensions to incrementally learn the new tasks, which can only process the task with a similar input. While the iKAN framework has multiple encoders aiming to receive different inputs and one fixed classifier, thus iKAN can incrementally learn the tasks across heterogeneous datasets.

The iKAN framework consists of three main components: encoders, feature redistribution layer, and classifier. 
The additional task identification layer is designed to recognize the datasets automatically according to the input dimension information, and the recognized task IDs $T_{id}$ provide external information to control the distribution of the features from the encoder by the feature redistribution layer. Besides, the global prediction results $C_{global}$ is determined by the task ID and within-task classification result $C_{local}$, and the output dimension of the classifier $N$ as \cref{eq: final_clf_num} presents. The output dimension of the classifier is fixed in the iKAN framework, which is equal to the maximal class number (13) among the six tested datasets.

\begin{equation}
\label{eq: final_clf_num}
    C_{global} = T_{id} \cdot N + C_{local}  
\end{equation}

\begin{figure}
\includegraphics[width=1.0\linewidth]{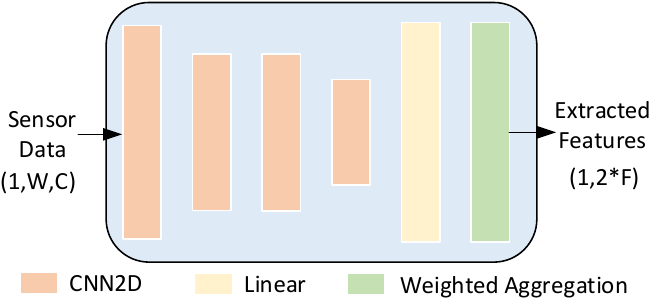}
\caption{The Architecture of Encoder (The input shape is (1, W, C), W: window size, C: sensor channel number, output shape is (1, 2*F), F: the number of out channels in CNN2D layers, where four CNN2D layers keep the same out channels)}
\label{fig: encoder}
\end{figure}

\cref{fig: encoder} shows the architecture of the encoder. There are four channel-wise convolutional layers with kernel size of (3,1)  extracting the features within individual sensor channels followed by one linear layer fusing the information across the sensor channels, and the cross-channel fused information is aggregated by the weighted aggregation layer \cite{zhou2022tinyhar}, by which the feature number can be further compressed and the output dimension is dependent from the input shape of the encoder. 
The encoder along with a two-linear-layer classifier is trained separately in different tasks, then the encoder is frozen when training the complete iKAN framework.
In the experiment, we kept the output channel number of the convolutional layers at 10, the encoder output 20 features. 
Therefore, the classifier can keep the same input dimension to process the data from heterogeneous datasets.

Since the core idea of the iKAN framework is to explore the local plasticity of the KAN to address the catastrophic forgetting challenge, a feature redistribution layer was designed to avoid the overlap of the feature representation from different tasks. In the feature redistribution layer, a normalization layer was applied to the extracted features, and the normalized features $F_{norm}$ are in the range from 0 to 1, then we scaled the magnitude of the normalized feature according to the number of the total tasks $N$ and its task ID $T_{id}$. 
Besides, the normalization operation can guide to set the grid range in the KAN, which is a hyperparameter that should be configured before training \cite{liu2024kan}. 
The redistributed features from different tasks can be calculated according to \cref{eq: redistribution}, where the $\beta$ is a hyperparameter to control the distance between the features from each task.

\begin{equation}
\label{eq: redistribution}
    F_{re} = \frac{T_{id}}{N} + \frac{F_{norm}}{N + \beta}
\end{equation}

\begin{figure*}[!t]
\footnotesize
\centering
\includegraphics[width=1.0\linewidth]{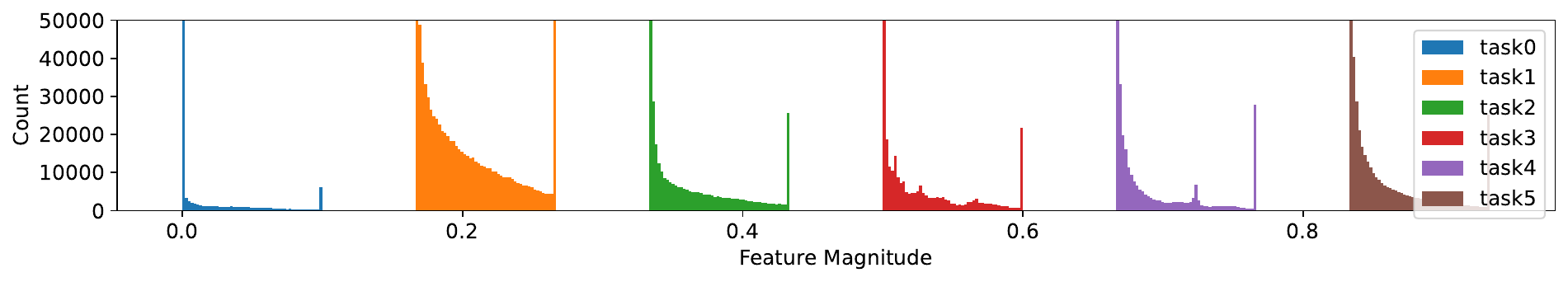}
\caption{Feature distribution result after the feature redistribution layer}
\label{fig: redistribution}
\end{figure*}

\cref{fig: redistribution} shows the redistribution of the extracted features from different tasks, where $\beta = 4$. It can be observed that there are non-overlapped features between each task. 

In the iKAN framework, rather than employing MLPs as the classifier, the KAN is chosen to mitigate catastrophic forgetting by utilizing the local plasticity of splines. Since spline bases are local, a sample will impact only a few nearby spline coefficients, leaving distant coefficients unaffected (which is beneficial as distant regions may contain information we wish to retain). In contrast, MLPs typically use global activations, where any local modification can spread uncontrollably to distant regions, potentially disrupting the stored information \cite{liu2024kan}.
In the iKAN framework, the classifier's architecture is always fixed. The input channel is equal to the number of features extracted from the encoder, and the number of output channels is configured as 13, which is the maximal class number of the tested datasets.
Besides, the performance of the KAN is closely related to its grid range and the grid number. In the experiment, we set the grid range from 0 to 1 due to the normalized features.
The grid number can not be determined according to the input information; too many or too few grids can lead to poor performance. 
Thus, the optimal KAN grid configurations for the selected HAR tasks were studied through empirical investigation.

\section{Evaluation}

To assess the performance of the proposed iKAN framework, we utilized six publicly available sensor-based HAR datasets featuring various sensor modalities and channels. Additionally, three types of metrics were chosen to evaluate the framework's ability to balance the model's plasticity and stability when handling incremental streaming data for both new and existing tasks.

\subsection{Dataset}
\cref{tab:datasets} shows detailed information about the six selected public HAR datasets.
It can be observed that the sensor modalities, location, and sampling rates are very different among these six datasets. Besides, the number of activity classes ranges from 6 to 13, and the window sizes were configured as different lengths, which is completely different from the existing works that tested the new tasks from new datasets with a similar input dimension.
To evaluate whether the framework can recognize different activities across the subjects, a leave-n-subjects-out cross-validation method was utilized. 
The number of subjects left out varies across datasets, which is configured based on the total number of subjects in each dataset.

\begin{table*}[ht]
\footnotesize
\centering
\caption{Datasets for test}
\label{tab:datasets}
\footnotesize
\begin{threeparttable}
\begin{tabular}{c l  l c l c c l} 
\toprule
ID&Dataset (Year)   & Classes (Num)              & Subjects & Sensors (Channels)& Sampling Rate & Window Size & Validation Method\\ \midrule
0&Opportunity (2013) \cite{chavarriaga2013opportunity}   & daily complex activities (12 $^1$) & 4 & 7 IMUs and 12 ACC (113)& 30 &90&Leave one subject out\\
1&PAMAP2 (2012) \cite{reiss2012pamap2}   & daily complex activities (13) & 8 & 3 IMUs and 1 HRS (39)&100&200&Leave two subjects out\\ 
2&MHealth (2014)\cite{banos2014mhealth} & daily and sports activities (12)& 10 & 3 IMUs and 2-lead ECG (21)&50&100&Leave two subjects out\\
3&WISDM (2011)\cite{kwapisz2011wisdm} & daily locomotion activities (6)  & 36       & 1 ACC (3)& 20 &80&Leave four subjects out\\
4&MotionSense (2018)\cite{malekzadeh2018motionsense}& daily locomotion (6) & 24 & 1 smartphone in the pocket (12)&50&200&Leave four subjects out\\
5&MM-Fit (2020)\cite{stromback2020mmfit}& sports activities(11) & 10 & 5 IMUs (24)&50&200&Leave two subjects out\\
\bottomrule
\end{tabular}
\begin{tablenotes}
\item[1] The original gesture class number is 18, we merged some activities, like all the activities about opening doors are merged into one class.
\end{tablenotes}

\end{threeparttable}
\end{table*}

\subsection{Evaluation Metrics}
In order to evaluate iKAN's ability to incrementally learn and adapt to new classes while retaining knowledge about previous classes, we compute several measures, including prediction Performance, Forgetting, and Intransigence Measures \cite{chaudhry2018riemannian}.

In the prediction performance measures, we refer to the two metrics in \cite{lu2024revisiting}, the Last performance (LP) and Average Incremental Performance (AIP). 
In this work, we specifically select the weighted F1 score to evaluate the prediction performance.
The LP, i.e.,$ P_{k} = \frac{1}{k}\sum^{k}_{i}{P_{i,k}}$, is the classification performance after the last task, reflecting the overall performance of all classes.
The AIP denotes the average of $ P_{k}$ over all tasks, thus $AIP = \frac{1}{N}\sum^{N}_{k}{P_{k}}$.
The higher the LP and AIP, the better the incremental learning performance.

Forgetting is a critical aspect of monitoring the efficiency of incremental learning, where a model is trained continuously on new data without having access to all the past data. The \textbf{Forgetting Measure} is to quantify how much the model has "forgotten" the knowledge from previous tasks as it learns new tasks.
The forgetting measure for a specific task $T_j$ after training on task $T_i$ is given by \cref{eq: forgetting}, where $P$ is the performance socre (weighted F1 Score).
\begin{equation}
\label{eq: forgetting}
    F(T_j, T_i) = \max_{k \in \{j, j+1, \ldots, i-1\}} P(T_k, T_j) - P(T_i, T_j)
\end{equation}
And the aggregate forgetting measure up to task $T_i$ is calculated according to $F_i = \frac{1}{i-1} \sum_{j=1}^{i-1} F(T_j, T_i)$. The aggregate forgetting measure is used in this work. 


The \textbf{Intransigence Measure} assesses the ability of a model in an incremental learning scenario to acquire new tasks relative to a baseline model trained from scratch on those tasks. This is calculated by comparing the performance of the incremental learning model with that of a reference model trained from scratch on each task and then averaging the performance differences. This measure is essential for evaluating the efficiency and adaptability of incremental learning models.
The intransigence measure for a specific task $T_i$ is calculated according to $I(T_i) = P_{\text{reference}}(T_i) - P_{\text{incremental}}(T_i)$. The aggregate intransigence measure across all tasks is given by: $I_{\text{average}} = \frac{1}{n} \sum_{i=1}^{n} I(T_i)$.

\subsection{Experiment Configuration}

In the experiment, the datasets were assigned to unique task IDs as shown in \cref{tab:datasets}. 
The iKAN framework was trained using a two-step training schedule: first, the encoder was trained, and then the classifier was trained while freezing the encoder.
In the first step, the encoder model connected an MLP classifier was trained by 100 epochs with an early stop configuration, whose patience was set as 30. 
The cross-entropy loss function and Adam optimizer with a learning rate of 0.1 and 512 batch size were used during the training process.
The training and validation datasets were selected randomly among the subjects following the leave-n-suject-out cross-validation rule as shown in \cref{tab:datasets}. 
In the second step, we froze the encoder and trained the KAN-based classifier.
The grid range of the KAN-based classifier was set from 0 to 1, as the magnitude of extracted features from each dataset was controlled in the range of 0 to 1 by the feature redistribution layer, as \cref{fig: redistribution} shows.
The order of piecewise polynomials was configured as 3, and we disabled the grid update during the training process.
The dataset splitting in the classifier training is the same as the encoder training.
The KAN-based classifier was trained for 30 epochs with the Adam optimizer with a learning rate of 0.1 and cross-entropy loss function.

\begin{figure*}[ht]
\footnotesize
\centering
     \begin{subfigure}[b]{0.33\textwidth}
         \centering
         \includegraphics[width=\textwidth]{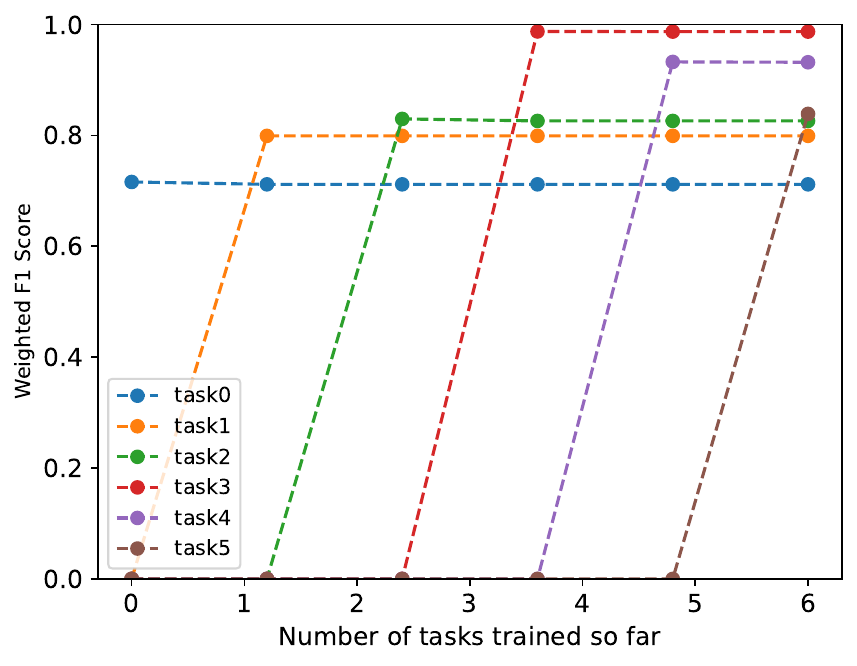}
         \caption{iKAN (Ours)}
         \label{fig:iKAN_result}
     \end{subfigure}
     \hfill
     \begin{subfigure}[b]{0.33\textwidth}
         \centering
         \includegraphics[width=\textwidth]{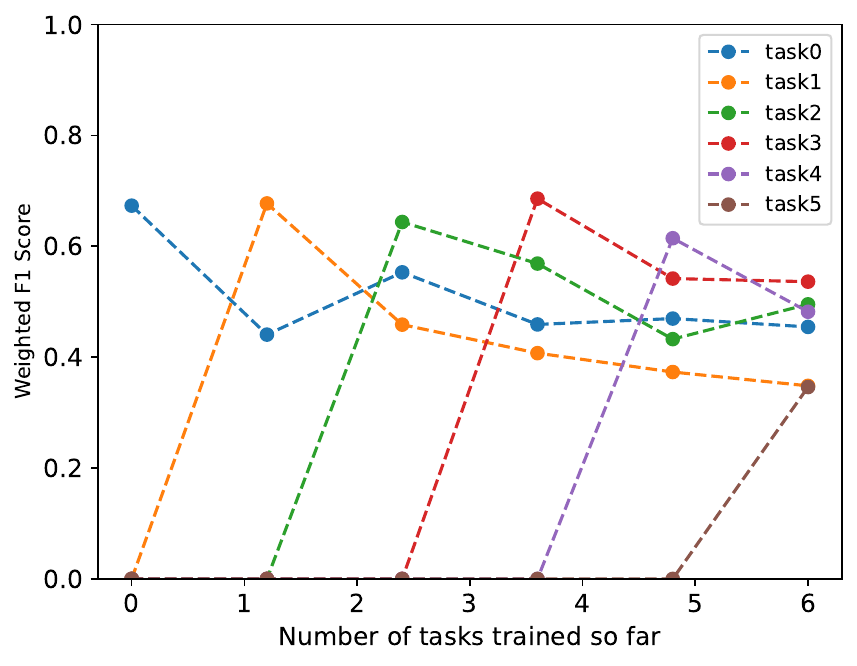}
         \caption{EWC method}
         \label{fig:EWC_result}
     \end{subfigure}
     \hfill
     \begin{subfigure}[b]{0.33\textwidth}
         \centering
         \includegraphics[width=\textwidth]{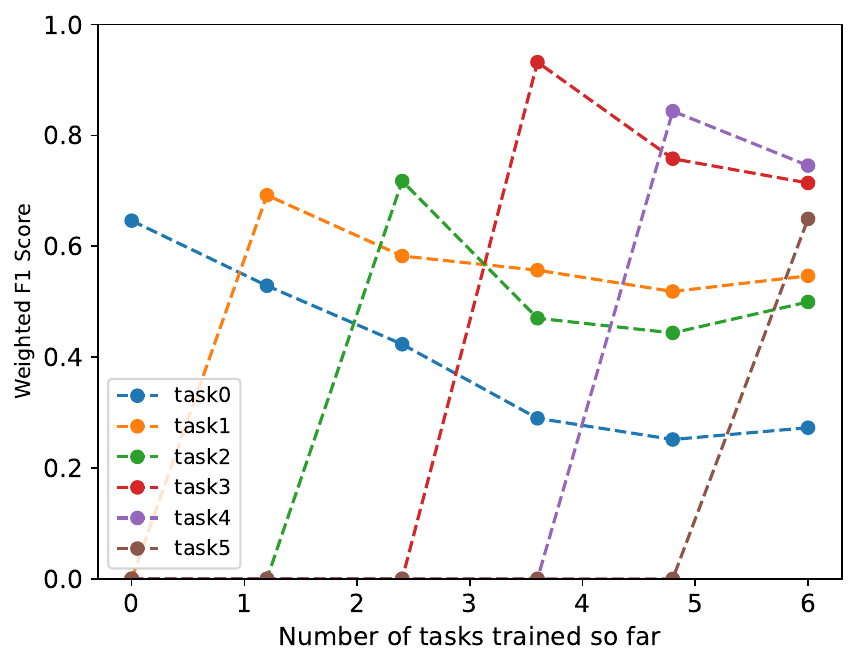}
         \caption{Replay method}
         \label{fig:replay_result}
     \end{subfigure}
    \caption{The performance of incremental learning by different methods (The grid number was set was 30 in iKAN, the lambda was configured as 80 in EWC, and store 200 samples of each class in the experience replay methods)}
    \label{fig:result-curve}
\end{figure*}

To evaluate the performance of the proposed iKAN framework comparatively, another two existing incremental learning methods, such as EWC and experience replay, were also tested while training the classifier with the frozen encoder. We refer to the model configuration of the two existing solutions in \cite{van2022three}.
In this experiment, the modification of the iKAN framework is as follows: the KAN-based classifier was replaced by the two-layer MLP-based classifier with 128 hidden states, and the feature redistribution layer was removed (we found the incremental learning performance of the MLP-based classifier followed by the feature redistribution is pretty bad in the experiment).
In addition, we also tested the effect of hyperparameters in the two existing solutions: the lambda in EWC, which controls the strength of the regularization term and determines how much importance is given to preserving the old weights while learning new tasks, and the memory size in experience replay methods.

\subsection{Results}

\cref{fig:result-curve} shows the weighted F1 score while incrementally learning the new tasks across different datasets(tasks).
It can be observed that the weighted F1 score of the old task remained almost the same as before in iKAN, as a result of the advantage of the local plasticity of the KAN, meaning that the learning of the new task will not affect the performance of the old tasks.
Compared to the existing two methods (EWC and replay solutions), where the weighted F1 score of the old task becomes worse when the model learns more new tasks, meaning that the memory of the old task continually decays during the process of learning new knowledge.
In addition, while the performance of learning the first task ($task0$) is similar among the three methods, there is an obvious performance degradation when learning the later tasks in EWC and replay methods compared to the iKAN. For example, the performance at learning $task3$ in EWC is much worse than the one in iKAN, as \cref{fig:iKAN_result} and \cref{fig:EWC_result} show. This means that the learning of the previous task could affect the performance of learning the next tasks in the EWC.
In the experiment, the performance of the replay methods shows a better performance than EWC, but the clear forgetting phenomenon still can be observed, especially the $task0$,  as \cref{fig:replay_result} shows.

\begin{table}
\small
  \caption{Result summary of incremental learning across heterogeneous datasets with different methods and configurations}
  \label{tab:result}
  \begin{threeparttable}
  \begin{tabular}{l c c c}
    \toprule
    Method & Configuration$^1$& LP & AIP \\
    \hline

    \hline
    \multirow{3}{*}{EWC}    &10     &41.7\% ±12.02\% & 46.73\% ± 2.57\%\\
                            &80    &44.37\% ± 4.95\% & 51.42\% ± 4.34\%\\
                            &100   &40.5\% ± 7.68\% & 46.99\% ± 4.97\%\\\hline
    \multirow{3}{*}{Replay} &10     &51.9\% ± 4.90\% & 53.05\% ± 0.72\%\\
                           &200      &57.1\% ± 1.85\% & 57.63\% ± 1.77\%\\
                           &800      &55.6\% ± 1.65\% & 59.92\% ± 2.92\%\\\hline
    \multirow{4}{*}{iKAN}&5         &18.3\% ± 2.52\% & 20.87\% ± 6.07\%\\
                          &15       &61.4\% ± 5.99\% & 62.16\% ± 3.24\%\\
                          &25 &83.04\% ± 0.61\% & 80.33\% ± 3.16\% \\
                          &30       &\textbf{84.9\% ± 0.43\%} & \textbf{81.34\% ± 3.87\%}\\

  \bottomrule
\end{tabular}
\begin{tablenotes}
\item[1] In the EWC method, we configured the hyperparameter Lambda with different values; In the Relay method, we configured the memory buffer size, the data represents the sample number of each class stored in the memory; In iKAN, the data represents the grid number.
\end{tablenotes}
\end{threeparttable}
\end{table}

\cref{tab:result} lists the last performance (LP) and average incremental performance (AIP) result of incremental learning across heterogeneous datasets with different methods and configurations.
The best result of the LP of 84.9 \% and the AIP of 81.34 \% was achieved by the proposed iKAN framework, which is around 30 \% and 20 \% higher than the replay method.
The EWC has the poorest performance among the three methods, whose LP and AIP are around 40 \% and 30 \% lower than iKAN's, respectively.
Overall, the iKAN demonstrated a remarkable advantage in incremental learning across heterogeneous datasets over the two other methods.

\begin{figure}
\includegraphics[width=1.0\linewidth]{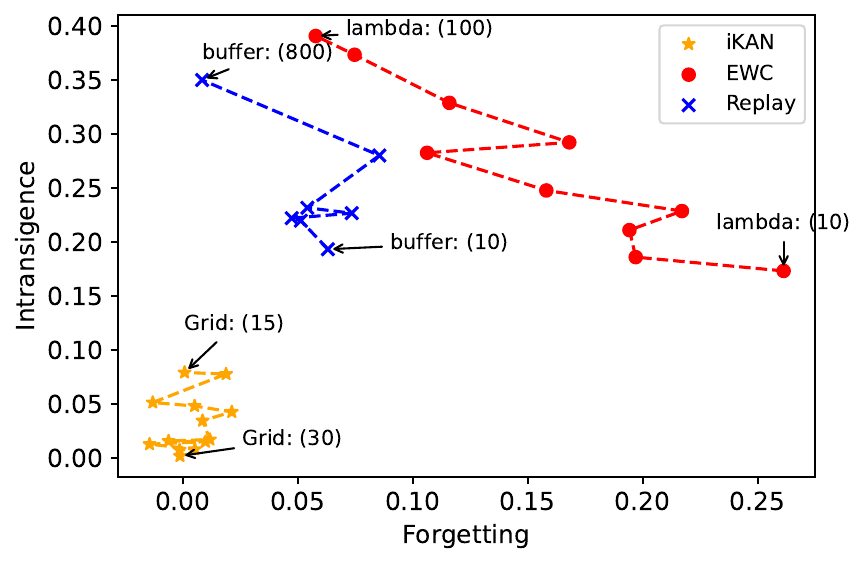}
\caption{Comparasion of the performance of the iKAN framework and existing methods in terms of forgetting and intransigence measure, the smallest result achieved by iKAN with 30 grid, whose forgetting measure is 0.0017, intransigence measure is 0.0012. (both the smaller the better)}
\vspace{-10pt}
\label{fig: FI}

\end{figure}

\cref{fig: FI} shows the result of the forgetting and intransigence measure.
The forgetting measure quantifies how much the model has "forgotten" the knowledge from previous tasks as it learns new tasks, while the intransigence measure is crucial for evaluating the efficiency and adaptability of incremental learning models.
The smaller, the better in both measurements.
It can be observed that the iKAN framework achieved the smallest result in both measurements, meaning that it can not only well remember previous knowledge but also easily learn new tasks.
In addition, the forgetting and intransigence measures in the EWC and replay are closely related to the hyperparameter, such as the lambda and memory buffer size.
Taken as a whole, the larger the size of the memory buffer, the better the model can remember the old knowledge, but it becomes more difficult to accept the new knowledge in the experience replay method. A similar phenomenon can be observed in EWC when configuring the hyperparameter Lambda.  Thus, it is difficult to keep both forgetting and intransigence measures smaller in the two existing methods. However, this issue can be well addressed by the proposed iKAN framework.


\begin{figure}[ht]
\footnotesize
\centering
     \begin{subfigure}[b]{0.45\textwidth}
         \centering
   
         \includegraphics[width=\textwidth]{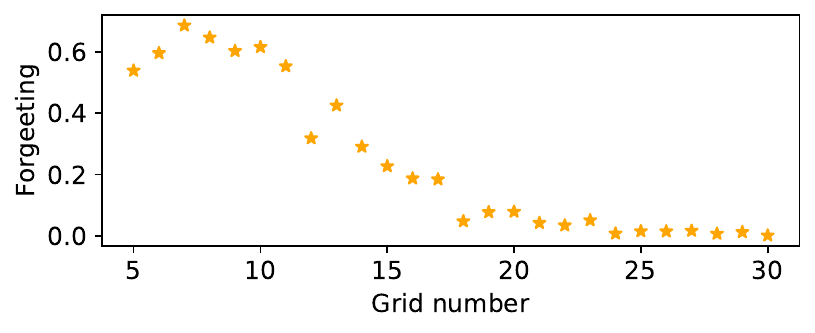}
        \caption{The forgetting measure of iKAN framework with different grid numbers}
        \label{fig: forgetting}
     \end{subfigure}
     \vfill
     \begin{subfigure}[b]{0.45\textwidth}
         \centering
         \includegraphics[width=\textwidth]{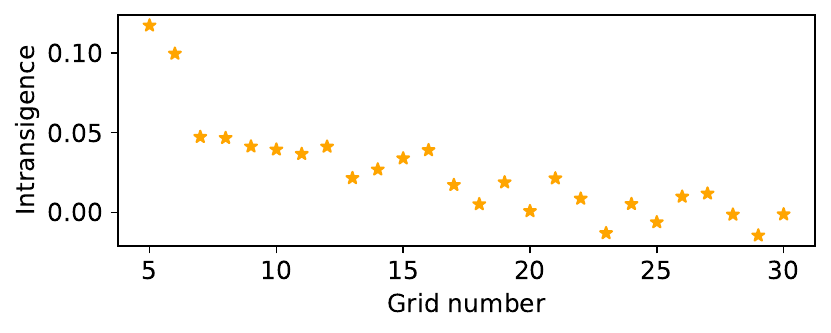}
        \caption{The intransigence measure of iKAN framework with different grid numbers}
        \label{fig: intransigence}
     \end{subfigure}
     \vspace{-10pt}
    \caption{The impact of grid number in KAN on the forgetting and intransigence measures  (the smaller the better)}
    
     \vspace{-10pt}
    \label{fig:grid-impact}
\end{figure}

\subsection{Discussion and Limitations}



Since the grid number of the KAN-based classifier in the proposed iKAN framework potentially impacts incremental learning performance, this study further investigates the grid's effect. Grid numbers ranging from 5 to 30, with a step size of 1, were chosen to test their effect on the forgetting and intransigence measures. \cref{fig: forgetting} and \cref{fig: intransigence} present the test results. It can be seen that both the forgetting and intransigence measures decrease as the grid number increases. The grid number's effect on the forgetting measures is more significant than on the intransigence measures; 
in other words, the incremental learning model's efficiency and adaptability are less dependent on the grid number. 
Additionally, the forgetting measure's value did not significantly decrease when the grid number exceeded 25. The grid is a crucial hyperparameter in the KAN, directly affecting performance and model size. Therefore, the method for selecting an appropriate grid number should be further explored in future work.

Although the iKAN framework demonstrated outstanding performance in incremental learning experiments across six public sensor-based HAR datasets, several limitations were observed in this study. Firstly, iKAN cannot be considered a fully class-incremental learning framework according to the definition of the three types of incremental learning in \cite{van2022three}, resulting from its task identification layer being restricted to recognizing tasks with different window dimensions; Otherwise, the task ID must be manually specified. Future research will investigate more advanced task identification methods to evolve iKAN into a fully class-incremental learning framework. Secondly, the impact of the hyperparameter $\beta$ from the feature redistribution layer on incremental learning performance has yet to be fully examined. Thirdly, the iKAN framework was primarily assessed using sensor-based datasets, and its performance on computer-vision HAR datasets remains to be evaluated in future work.

\section{Conclusion}
In this work, we introduced iKAN, an innovative incremental learning framework enhanced by KAN, enabling neural networks to continuously learn new tasks across heterogeneous HAR datasets. In the iKAN framework, we designed the feature redistribution layer to leverage the local plasticity and global stability benefits of KAN to mitigate catastrophic forgetting in IL. The IL performance of the iKAN framework was evaluated using six public HAR datasets, achieving a last performance of 84.9\% (weighted F1 score) and an average incremental performance of 81.34\% (weighted F1 score), significantly outperforming two existing methods, namely EWC and experience replay. 
Different from other IL methods for HAR, which are validated within each individual dataset and mask out the `new tasks', we test across datasets in a single validation process and leave out entire datasets as the new tasks.
Additionally, we examined the impact of the grid number in the iKAN framework on forgetting and intransigence metrics. The result confirmed the effective performance of the proposed iKAN framework.



\bibliographystyle{ACM-Reference-Format}
\bibliography{sample-base}
\end{document}